\documentclass[letterpaper]{article} 
\usepackage{aaai25}  
\usepackage{times}  
\usepackage{helvet}  
\usepackage{courier}  
\usepackage[hyphens]{url}  
\usepackage{graphicx} 
\usepackage{natbib}  
\usepackage{caption} 
\usepackage{algorithm}
\usepackage{algorithmic}

\usepackage{amsmath} 
\usepackage{amssymb} 
\usepackage{amsfonts}
\usepackage{bm} 
\usepackage{subfigure} 
\usepackage{multirow} 
\usepackage{booktabs}
\usepackage{graphicx}
\usepackage{utfsym}
\usepackage{makecell}
\usepackage{colortbl}

\usepackage{newfloat}
\usepackage{listings}
\DeclareCaptionStyle{ruled}{labelfont=normalfont,labelsep=colon,strut=off} 
\floatstyle{ruled}
\newfloat{listing}{tb}{lst}{}
\floatname{listing}{Listing}
\title{Sample-aware Adaptive Structured Pruning for Large Language Models}
\author{
   Jun Kong, Xinge Ma, Jin Wang\thanks{Corresponding author.}, Xuejie Zhang\\
}
\affiliations{
     
    School of Information Science and Engineering\\

    Yunnan University\\
    Kunming, China\\
    \{kongjun, maxinge\}@mail.ynu.edu.cn, \{wangjin, xjzhang\}@ynu.edu.cn
}

\usepackage{bibentry}

\begin{document}

\maketitle

\begin{abstract}
Large language models (LLMs) have achieved outstanding performance in natural language processing, but enormous model sizes and high computational costs limit their practical deployment. Structured pruning can effectively reduce the resource demands for deployment by removing redundant model parameters. However, the randomly selected calibration data and fixed single importance estimation metrics in existing structured pruning methods lead to degraded performance of pruned models. This study introduces AdaPruner, a sample-aware adaptive structured pruning framework for LLMs, aiming to optimize the calibration data and importance estimation metrics in the structured pruning process. Specifically, AdaPruner effectively removes redundant parameters from LLMs by constructing a structured pruning solution space and then employing Bayesian optimization to adaptively search for the optimal calibration data and importance estimation metrics. Experimental results show that the AdaPruner outperforms existing structured pruning methods on a family of LLMs with varying pruning ratios, demonstrating its applicability and robustness. Remarkably, at a 20\% pruning ratio, the model pruned with AdaPruner maintains 97\% of the performance of the unpruned model. 
\end{abstract}


\begin{links}
    \link{Code}{https://github.com/JunKong5/AdaPruner}
\end{links}

\section{Introduction}
Large language models (LLMs), such as GPT-3~\cite{Brown2020}, OPT~\cite{Zhang2022}, LLaMA~\cite{Touvron2023} and Vicuna~\cite{chiang2023vicuna}, have demonstrated significant accomplishments in the realm of natural language processing~\cite{Wei2022,Wu2020}. Nevertheless, their exceptional capabilities are coupled with substantial model sizes and elevated computational expenses. Furthermore, owing to the scaling law~\cite{Hoffmann2022,Kaplan2020}, LLMs tend to enhance model performance by progressively augmenting model parameters. Regrettably, larger model sizes entail heightened consumption of computational resources, presenting a notable obstacle to their practical deployment, particularly in settings with limited resources.

Structured pruning~\cite{Xia2024} emerges as a pivotal technique for mitigating resource demands in the deployment of LLMs. In comparison to other strategies~\cite{Zhu2023} like unstructured pruning~\cite{Yin2023,Frantar2023,Jaiswal2023}, model quantization~\cite{Liu2023,Xiao2023}, and knowledge distillation~\cite{Gu2024,Yuan2023,Hsieh2023}, structured pruning not only provides a practical and hardware-independent solution but also offers an effective approach to streamline LLMs implementation on devices with limited computational resources. Its unique ability to selectively remove redundant model parameters while maintaining model integrity positions structured pruning as a cornerstone in optimizing the efficiency of LLMs deployment.

However, existing work on structured pruning commonly employs Taylor expansion as the metric for estimating the importance of structures. These methods hinge on a localized approximation of the loss function, necessitating additional calibration data for gradient information computation. As a result, the precision of the gradient is directly tied to the calibration data’s quality, thereby influencing both the Taylor expansion approximation and the pruning decision. Subpar calibration data and unsuitable importance estimation metrics can lead to substantial performance degradation in pruned models. Thus, ensuring the quality of the calibration data and employing appropriate importance estimation metrics are crucial for effective structured pruning.

\begin{figure}[!t]
	\centering
	\subfigure[Calibration data]{
		\label{fig:Figure_1a}
		\includegraphics[scale=0.237]{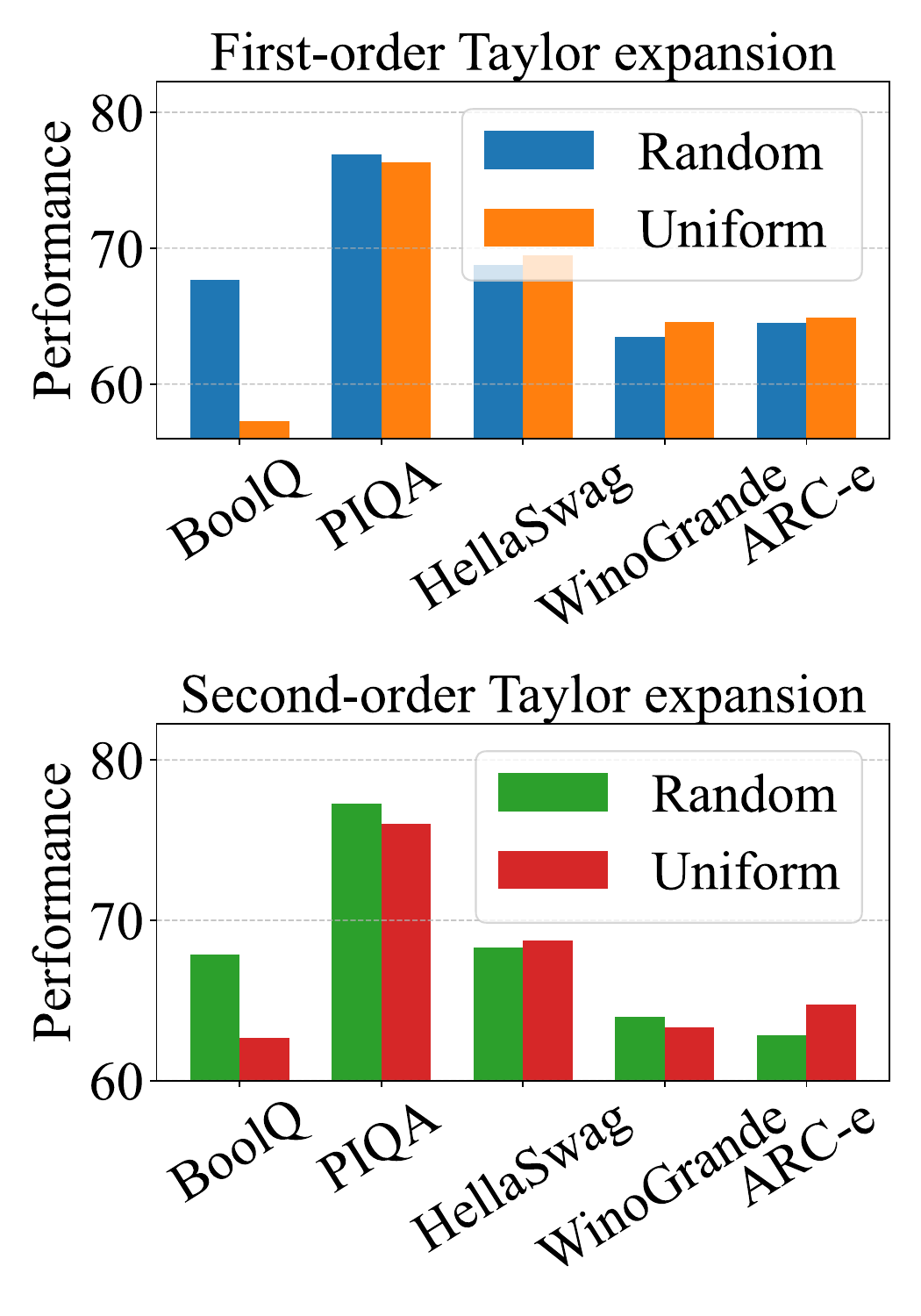}
	}
	\subfigure[Importance estimation]{
		\label{fig:Figure_1b}
		\includegraphics[scale=0.237]{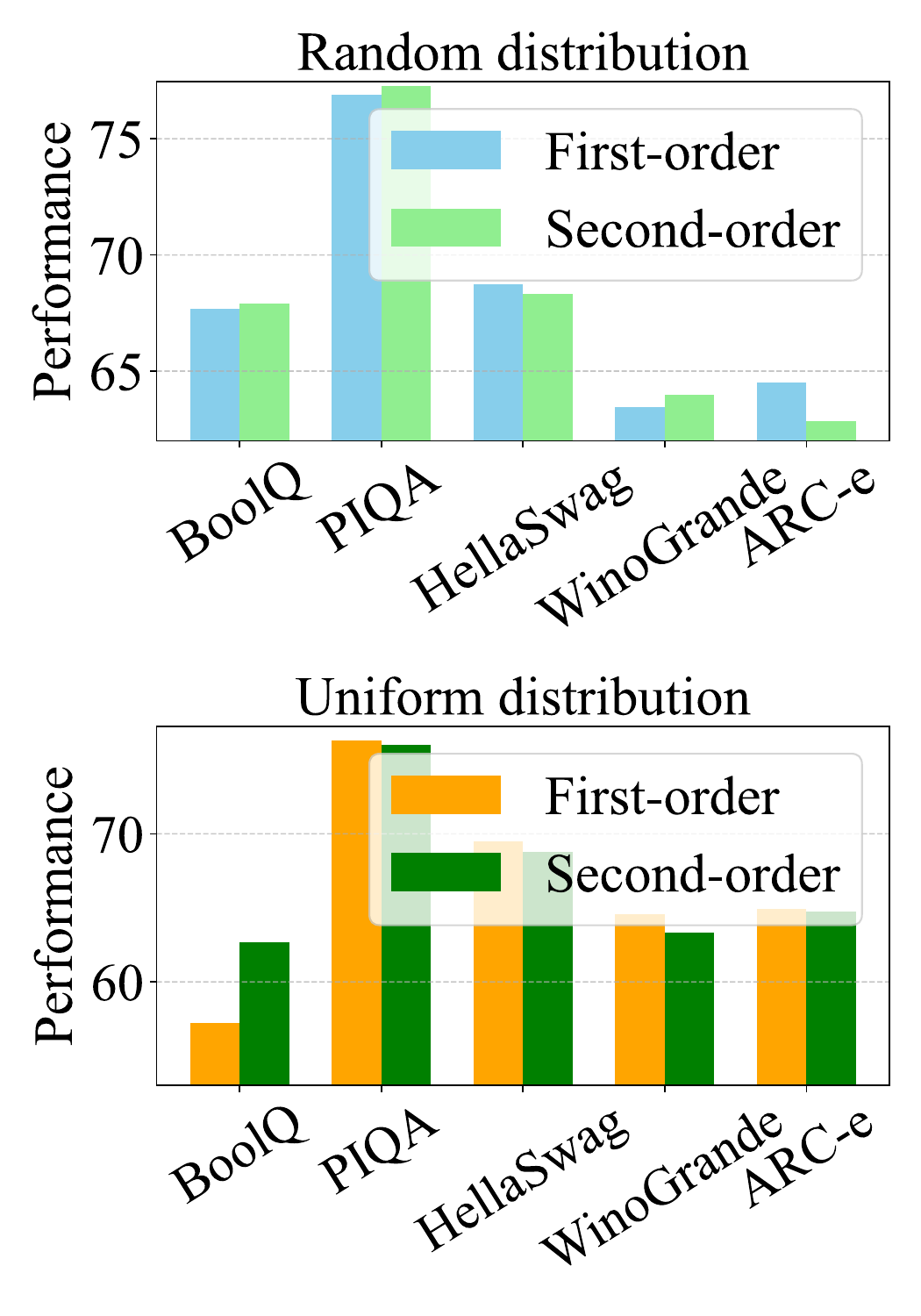}
	}
	\caption{The impact of calibration data and importance estimation metrics.}
	\label{fig:Figure_1}
\end{figure}

Hence, the Taylor expansion-based methodologies for structured pruning LLMs pose two notable challenges. Acquiring high-quality calibration data covering diverse parameters and sample spaces is crucial for accurate gradient computation. It aids in identifying parameters effectively during pruning, reducing computational overhead while preserving generalization. However, randomly selected calibration data lacks representativeness and diversity, leading to inaccurate gradient information and flawed pruning strategies. A preliminary experiment in Figure~\ref{fig:Figure_1a} examines the impact of different calibration data distributions on performance. It illustrates that uniformly distributed calibration data may degrade performance in some tasks and improve it in others compared to randomly distributed data. In addition, it highlights the significant influence of calibration data distributions on performance. Randomly selecting calibration data is considered suboptimal.

Further, prevailing structured pruning approaches typically employ predetermined heuristic metrics to appraise sub-structure significance. However, these metrics often fall short of comprehensively assessing the model parameters. Furthermore, their disregard for alignment with calibration data results in inaccurate estimation of model weight importance, consequently undermining model performance. This deficiency can be ascribed to the reliance on model parameter importance estimation on gradient information, which is sensitive to variations in calibration data. Consequently, disparate calibration datasets lead to varying gradient magnitudes, causing parameters previously deemed crucial to lose importance with new calibration data and vice versa. As depicted in Figure~\ref{fig:Figure_1b}, distinct importance estimation metrics exhibit varying sensitivity to calibration data. Thus, selecting compatible importance estimation metrics tailored to high-quality calibration data is imperative to mitigate performance degradation.

This study introduces AdaPruner, a sample-aware adaptive structured pruning framework for LLMs. AdaPruner aims to optimize both the calibration data and importance estimation simultaneously. Firstly, AdaPruner creates a structured pruning space containing a calibration data subspace and an importance estimation metrics subspace that are closely linked. Secondly, AdaPruner uses Bayesian optimization to search for high-quality calibration data and importance estimation metrics within this space. Lastly, AdaPruner conducts efficient pruning based on the obtained data and metrics, removing unnecessary structures from LLMs. AdaPruner streamlines the tedious manual design process and ensures stable performance by enabling the transfer of calibration data and importance estimation metrics to different pruning ratios and models.

The contributions can be summarized as follows:
\begin{itemize}
	\item AdaPruner introduces a sample-aware approach to structured pruning, delineating interconnected calibration data and importance estimation metrics subspaces, enhancing their utilization in the pruning process.
	\item AdaPruner facilitates rapid and effective pruning by utilizing acquired calibration data and importance estimation metrics to eliminate redundant LLMs structures, reducing manual effort and ensuring consistent performance across different pruning rates and models.
	\item We conduct extensive experiments on a variety of language benchmarks. The AdaPruner outperforms the existing methods on the LLaMA series models, achieving superior average performance over the LLM-Pruner by 1.37\%. The experimental results demonstrate the effectiveness of the proposed AdaPruner.
\end{itemize}

\section{Sample-aware Adaptive Structured Pruning}
\begin{figure*}[!t]
	\centering
	\includegraphics[width=\linewidth]{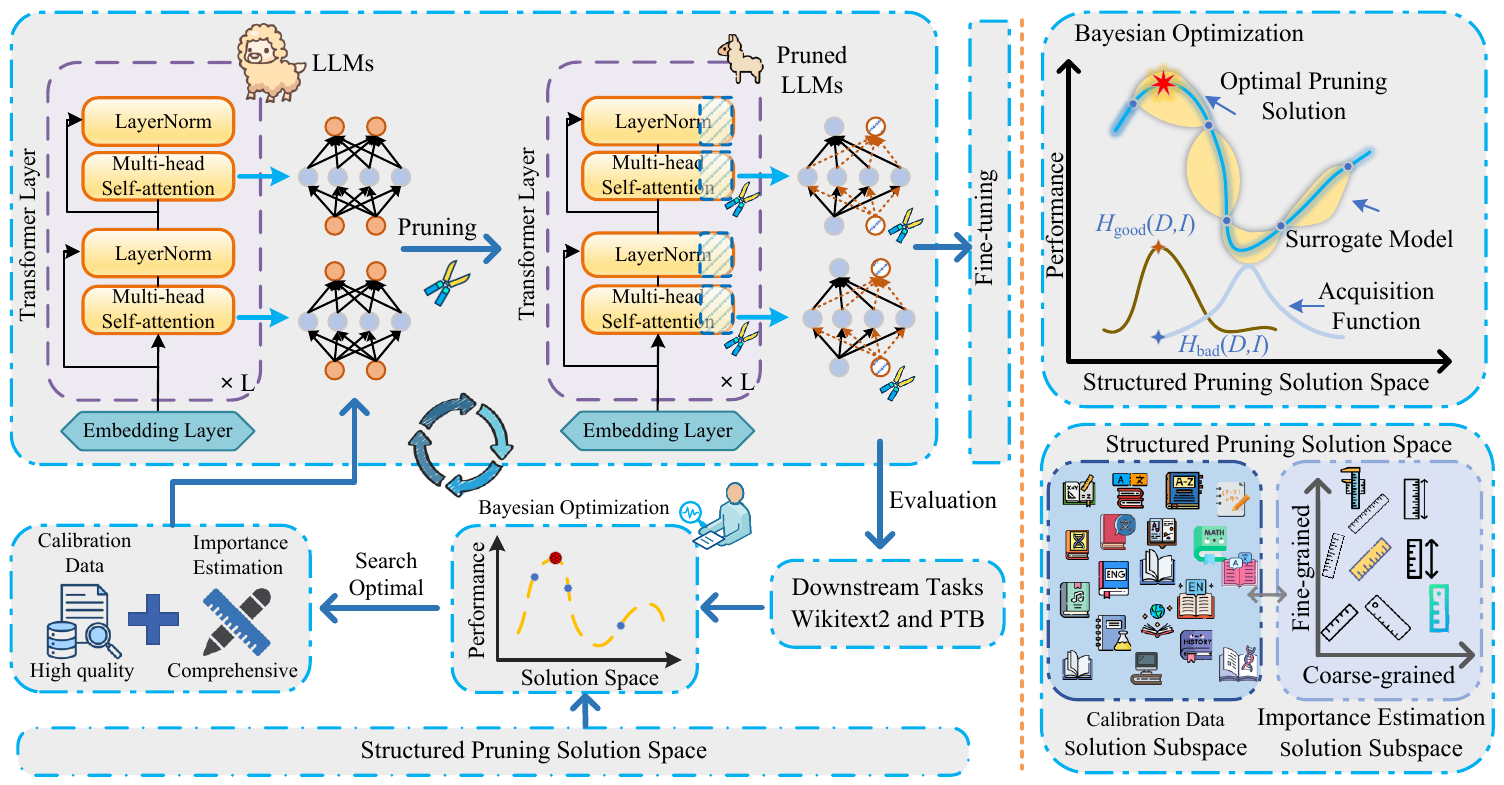}
	\caption{Overview of the AdaPruner Framework.} 
	\label{fig:Figure_2} 
\end{figure*}

This section introduces AdaPruner, a sample-ware adaptive structured pruning framework. As shown in Figure~\ref{fig:Figure_2}, AdaPruner aims to adaptively obtain the optimal calibration data and importance estimation metrics through Bayesian optimization to improve the performance of pruned LLMs.

\subsection{Structured Pruning for LLMs}
First, according to LLM-Pruner~\cite{Ma2023}, structured pruning considers the dependencies between structures. Therefore, the dependency automatically identifies and groups coupling structures in LLMs. 

Second, structured pruning is performed based on Taylor expansion. Specifically, given a calibration dataset $D = \{ {x_i}\} _{i = 1}^N$, where $N$ is the number of samples contained in the calibration dataset, and the subsequent token prediction loss of the LLMs parameterized by $\theta$ on the calibration dataset is:
\begin{equation}
    L(\theta ,D) = \frac{1}{N}\sum\limits_{i - 1}^N {F(\theta ,{x_i})}
\end{equation}
where $F(\theta ,{x_i})$ represents the subsequent token prediction loss of the sample $x_i$ on the LLM.

The importance of $\theta_i$ can be quantified by measuring the impact of its removal on the loss. Structured pruning methods typically estimate the effect of removing parameters through Taylor expansion as follows:
\begin{equation}\label{eq:Equation_2}
    \begin{array}{l}
    {I_{{\theta _i}}} = |L({\theta _i},D) - L({\theta _i} = 0,D)|\\
    \;\;\;\;{\rm{    }} \approx |\frac{{\partial L({\theta _i},D)}}{{\partial {\theta _i}}}{\theta _i} - \frac{1}{2}{\theta _i}^ \top H{\theta _i} + {\rm O}(\parallel {\theta _i}{\parallel ^3})|
    \end{array}
\end{equation}
where $L({\theta _i} = 0,D)$ denotes removing $\theta_i$ from the LLM, $H$ denotes the Hessian matrix, and ${\rm O}(\parallel {\theta _i}{\parallel ^3})$ denotes the residual term that can be ignored in the calculation. Then, the importance of the structural group is obtained by aggregating the importance of the structural parameters within the group, i.e., $I(G) = \mathop {Agg}\limits_{i = 1}^M I({\theta _i})$. Here, $Agg$ denotes aggregation functions such as summation, product, maximum, and last only. $M$ denotes the number of structures in the group. Finally, the groups with lower importance scores are removed based on a predefined pruning ratio.

\subsection{Solution Space for Structured Pruning}
We jointly design a structured pruning solution space consisting of the calibration data subspace and the importance estimation metrics subspace.
\paragraph{Subspace of calibration data.} Due to the restricted access to the pre-training datasets, existing studies usually sample public datasets and randomly select a small portion of samples from them as the calibration data. Following LLM-Pruner, we construct the subspace of the calibration data using samples from the BookCorpus dataset, as shown in Figure~\ref{fig:Figure_2}. Meanwhile, the subspace solution of the calibration data is effectively reduced by filtering out the short texts with little information, accelerating the search process.
\paragraph{Subspace of importance estimation metrics.} To comprehensively assess the importance of LLMs parameters, the importance estimation subspace comprises different granularity importance estimation metrics based on Taylor expansion, i.e., coarse-grained weight vectors and fine-grained weight element importance estimation metrics. The coarse-grained metric, as shown in Eq.~\eqref{eq:Equation_2}, is used to estimate the importance of the overall structure weights. The fine-grained metric is introduced as follows:
\begin{equation}
    \begin{array}{l}
    {I_{\theta _i^k}} = |L(\theta _i^k,D) - L(\theta _i^k = 0,D)|\\
    \;\;\;\;\;{\rm{     }} \approx |\frac{{\partial L(\theta _i^k,D)}}{{\partial \theta _i^k}}\theta _i^k - \frac{1}{2}{\theta _i^k }^\top{H_{kk}}\theta _i^k + {\rm O}(\parallel \theta _i^k{\parallel ^3})|
    \end{array}
\end{equation}
where $k$ denotes the index of the parameter. Due to the Hessian matrix ${H_{kk}}$ can be approximated by the Fisher in-formation matrix~\cite{Kwon2022}, It can be redefined as:
\begin{equation}\label{eq:Equation_4}
    \begin{array}{l}
    {I_{\theta _i^k}} = |L(\theta _i^k,D) - L(\theta _i^k = 0,D)|\\
    \;\;\;\;\;{\rm{     }} \approx |\frac{{\partial L(\theta _i^k,D)}}{{\partial \theta _i^k}}\theta _i^k - \frac{1}{2}{\sum\limits_{j = 1}^N {\left( {\frac{{\partial L(\theta _i^k,D)}}{{\partial \theta _i^k}}\theta _i^k} \right)} ^2} + {\rm O}(\parallel \theta _i^k{\parallel ^3})|
    \end{array}
\end{equation}

The granularity of the weight vector is too coarse, while the granularity of the weight element is too fine. Therefore, balancing weight vector and weight element metrics is necessary. Therefore, the importance estimation metrics subspace is defined as follows:
\begin{equation}
    {I_{{\theta _i},}}_{\theta _i^k} = {\alpha _1}{I_{{\theta _i}}} + {\alpha _2}{I_{\theta _i^k}}
\end{equation}
where $\alpha_1$ and $\alpha_2$ are equilibrium coefficients for coarse- and fine-grained importance estimations, respectively. However, there is a significant order of magnitude difference between the manually designed first- and second-order gradients. We further extend the optimization of the alignment factor.

\subsection{Adaptive Pruning via Bayesian Optimization}
The effects of calibration data and importance estimation metrics on the performance of pruned models are interrelated. Independently optimizing either element may lead to a locally optimal solution. Therefore, AdaPruner uses Bayesian optimization to adaptively obtain optimal calibration data and importance estimation metrics in the structured pruning solution space. Specifically, the pruned model $f(\theta ,D,I,\lambda )$ is evaluated with the dataset ${D_E}$, e.g., using the perplexity of the evaluation dataset as an evaluation metric. Thus, the optimization objective of LLMs pruning can be formulated as follows:
\begin{equation}\label{eq:Equation_6}
    H = \mathop {\min }\limits_{D,I \in Z} h(f(\theta ,D,I,\lambda ),{D_E})
\end{equation}
where $h( \cdot )$ denotes the evaluation metric, $\lambda$ denotes the pruning ratio, and $D$ and $I$ denote the calibration data and importance estimation metrics to search, respectively. The pruned model without fine-tuning is used to evaluate the performance on a smaller evaluation set, such as Wik-iText2 and PTB, to minimize the evaluation cost. Their average perplexity is utilized as a metric. Lower perplexity scores indicate superior language modeling capabilities, reflecting a more effective pruning metric regarding mod-el performance.

Since $h(f(\theta ,D,I,\lambda ),{D_E})$ is a black-box function, AdaPruner utilizes Bayesian Optimization with Tree-Structured Parzen Estimator (BO-TPE) to globally optimize Eq.~\eqref{eq:Equation_6} to obtain the optimal calibration data and importance estimation metrics, as shown in Figure~\ref{fig:Figure_2}. BO-TPE is an iterative process designed to optimize the objective function $h(f(\theta ,D,I,\lambda ),{D_E})$ from the structured pruning solution space $Z$. It adaptively obtains the optimal calibration data and importance estimation metrics. It uses a probabilistic surrogate model $M$ to estimate the objective function and updates the posterior estimate of $h(f(\theta ,D,I,\lambda ),{D_E})$ based on each search step's results. Specifically, given the previous $t$-1  search results $\{ ({D_1},{I_1}),...,({D_{t - 1}},{I_{t - 1}})\}$ and their evaluation results $H = [H({D_1},{I_1}),...,H({D_{t - 1}},{I_{t - 1}})]$, we iteratively update the surrogate model $M$ to estimate $h(f(\theta ,D,I,\lambda ),{D_E})$. In each iteration $t$, to determine the next pruning solution $({D_t},{I_t})$, we use the expected improvement (EI) acquisition function to compute the probability of improvement below a threshold ${H^*}$ given the observations   and $\{ ({D_1},{I_1}),...,({D_{t - 1}},{I_{t - 1}})\}$, as follows:
\begin{equation}
    E{I_{{H^ * }(D,I)}} = \int_{ - \infty }^{{H^ * }} {({H^ * } - H)p(H|(D,I))} dH
\end{equation}

Subsequently, AdaPruner stores $(D,I)$ and $H(D,I)$ into the search history and fits a new agent model based on the updated record $M$. At the end of the loop, AdaPruner outputs the globally optimal $(D,I)$. Specifically, instead of directly modeling, BO-TPE models $p(H|(D,I))$ by decomposing it into $p((D,I)|H)$ and $p(D,I)$ using Bayes theorem, as follows: 
\begin{equation}\label{eq:Equation_8}
    p(H|(D,I)) = \frac{{p((D,I)|H)p(H)}}{{p(D,I)}}
\end{equation}
\begin{equation}\label{eq:Equation_9}
    p((D,I)|H) = \left\{ {\begin{array}{*{20}{c}}
{{H_{good}}(D,I),{\rm{  if\;}}H <  {{H}^*}}\\
{{H_{bad}}(D,I),{\rm{  if\;}}H \ge {{H}^*}}
\end{array}} \right.
\end{equation}
    \begin{equation}\label{eq:Equation_10}
        \begin{array}{l}
    p(D,I) = \int_\mathbb{R} {p((D,I)|H)p(H)d} {H}\\
    \;\;\;\;\;\;\;\;\;\;\;\;{\rm{             = }}\gamma {H_{good}}(D,I) + (1 - \gamma ){H_{{\rm{bad}}}}(D,I)
    \end{array}
\end{equation}

The BO-TPE constructs different $p((D,I)|H)$ on different sides of the threshold, where ${H_{good}}(D,I)$ is the excellent density formed by searching into the pruning setup $(D,I)$ to ensure that the model's perplexity is below the threshold ${H^ * }$, and ${H_{good}}(D,I)$ is the lousy density formed by the residuals $(D,I)$. The threshold ${H^ * }$ is determined by the quantile $\gamma $ of the performance of the searched pruned model. Combining Eq.~\eqref{eq:Equation_8}, \eqref{eq:Equation_9}, and \eqref{eq:Equation_10}, the final EI of BO-TPE is expressed as:
\begin{equation}
    \begin{array}{l}
E{I_{{H^ * }(D,I)}} = \int_{ - \infty }^{{H^ * }} {({H^ * } - H)p(H|(D,I))} dH\\
\;\;\;\;\;\;{\rm{               = }}\int_{ - \infty }^{{H^ * }} {({H^ * } - H)\frac{{p((D,I)|H)p(H)}}{{p(D,I)}}dH} \\
\;\;\;\;\;\;{\rm{               = }}\int_{ - \infty }^{{H^ * }} {({H^ * } - H)\frac{{{H_{good}}(D,I)p(H)}}{{\gamma {H_{good}}(D,I) + (1 - \gamma ){H_{bad}}(D,I)}}dH} \\
\;\;\;\;\;\;{\rm{               = }}\frac{{\int_{ - \infty }^{{H^ * }} {({H^ * } - H)p(H)} dH}}{{\gamma  + (1 - \gamma )\frac{{{H_{bad}}(D,I)}}{{{H_{good}}(D,I)}}}} \propto {\left( {\gamma  + \frac{{{H_{bad}}(D,I)}}{{{H_{good}}(D,I)}}(1 - \gamma )} \right)^{ - 1}}
\end{array}
\end{equation}

This shows that maximizing EI is proportional to the ratio of maximizing $\frac{{{H_{good}}(D,I)}}{{{H_{bad}}(D,I)}}$. The ultimate goal is to maximize the probability of ${H_{good}}(D,I)$ while decreasing the probability of ${H_{bad}}(D,I)$. After obtaining the optimal $D$ and $I$, structured pruning is performed on the LLMs.

To improve the performance of the pruned model, we employ the Low-Rank Adaptation (LoRA)~\cite{Hu2022} to fine-tune the model on the Stanford Alpaca~\cite{taori2023stanford} dataset. Specifically, LoRA adds two low-rank matrices to the original weight matrix, avoiding the need for full-parameter fine-tuning, as follows:
\begin{equation}
    \theta  = \theta 'x + \Delta \theta x = \theta 'x + BAx
\end{equation}
where $A$ and $B$ are two learnable low-rank matrices.

\begin{table*}
\centering
\setlength{\extrarowheight}{0pt}
\addtolength{\extrarowheight}{\aboverulesep}
\addtolength{\extrarowheight}{\belowrulesep}
\setlength{\aboverulesep}{0pt}
\setlength{\belowrulesep}{0pt}

\resizebox{1\linewidth}{!}{
\begin{tabular}{cc!{\vrule width \lightrulewidth}cc!{\vrule width \lightrulewidth}ccccccc} 
\toprule
Pruning ratio                                                                 & Method        & WikiText2$\downarrow$                              & PTB$\downarrow$                                    & BoolQ$\uparrow$                                    & PIQA$\uparrow$                                     & HellaSwag$\uparrow$                                & WinoGrande$\uparrow$                               & ARC-e$\uparrow$                                    & ARC-c$\uparrow$                                    & OBQA$\uparrow$                             \\ 
\midrule
Ratio=0\%                                                                     & LLaMA-7B      & 12.62                                              & 22.14                                              & 73.18                                              & 78.35                                              & 72.99                                              & 67.01                                              & 67.45                                              & 41.38                                              & 42.40                                      \\ 
\midrule
\multirow{6}{*}{\begin{tabular}[c]{@{}c@{}}Ratio=20\%\\w/o tune\end{tabular}} & Magnitude     & 582.41                                             & 1022.1                                             & 59.66                                              & 58.00                                              & 37.04                                              & 52.41                                              & 33.12                                              & 28.58                                              & 29.80                                      \\
                                                                              & Vector        & 22.28                                              & 41.78                                              & 61.44                                              & 71.71                                              & 57.27                                              & 54.22                                              & 55.77                                              & 33.96                                              & 38.40                                      \\
                                                                              & LLM-Pruner-E1 & 19.09                                              & 34.21                                              & 57.06                                              & 75.68                                              & 66.80                                              & 59.83                                              & 60.94                                              & 36.52                                              & 40.00                                      \\
                                                                              & LLM-Pruner-E2 & 19.77                                              & 36.66                                              & 59.39                                              & 75.57                                              & 65.34                                              & 61.33                                              & 59.18                                              & 37.12                                              & 39.80                                      \\
                                                                              & MoreauPruner  & 18.61                                              & 32.92                                              & 55.44                                              & 76.17                                              & 66.47                                              & 63.61                                              & 61.53                                              & 37.80                                              & \textbf{40.60}                             \\
                                                                              & AdaPruner     & {\cellcolor[rgb]{0.906,0.969,0.973}}\textbf{17.72} & {\cellcolor[rgb]{0.906,0.969,0.973}}\textbf{31.10} & {\cellcolor[rgb]{0.906,0.969,0.973}}\textbf{62.39} & {\cellcolor[rgb]{0.906,0.969,0.973}}\textbf{76.33} & {\cellcolor[rgb]{0.906,0.969,0.973}}\textbf{68.03} & {\cellcolor[rgb]{0.906,0.969,0.973}}\textbf{63.64} & {\cellcolor[rgb]{0.906,0.969,0.973}}\textbf{63.64} & {\cellcolor[rgb]{0.906,0.969,0.973}}\textbf{38.65} & {\cellcolor[rgb]{0.906,0.969,0.973}}40.40  \\ 
\midrule
\multirow{7}{*}{\begin{tabular}[c]{@{}c@{}}Ratio=20\%\\w/ tune\end{tabular}}  & Magnitude     & 21.78                                              & 38.64                                              & 61.89                                              & 70.81                                              & 58.34                                              & 56.87                                              & 54.87                                              & 34.02                                              & 38.40                                      \\
                                                                              & Vector        & 18.84                                              & 33.05                                              & 65.75                                              & 74.70                                              & 64.52                                              & 59.35                                              & 60.65                                              & 36.26                                              & 39.40                                      \\
                                                                              & LLM-Pruner-E1 & 17.58                                              & 30.11                                              & 64.62                                              & 77.20                                              & 68.80                                              & 63.14                                              & 64.31                                              & 36.77                                              & 39.80                                      \\
                                                                              & LLM-Pruner-E2 & 17.37                                              & 30.39                                              & 69.54                                              & 76.44                                              & 68.11                                              & 65.11                                              & 63.43                                              & 37.88                                              & 40.00                                      \\
                                                                              & LoRAShear     & -                                                  & -                                                  & 70.17                                              & 76.89                                              & 68.69                                              & \textbf{65.83}                                     & 54.11                                              & 38.77                                              & 39.97                                      \\
                                                                              & MoreauPruner  & 17.01                                              & 30.27                                              & 66.61                                              & 77.04                                              & 68.32                                              & 65.59                                              & 65.57                                              & 38.40                                              & \textbf{41.20}                             \\
                                                                              & AdaPruner     & {\cellcolor[rgb]{0.906,0.969,0.973}}\textbf{16.75} & {\cellcolor[rgb]{0.906,0.969,0.973}}\textbf{28.71} & {\cellcolor[rgb]{0.906,0.969,0.973}}\textbf{70.34} & {\cellcolor[rgb]{0.906,0.969,0.973}}\textbf{77.69} & {\cellcolor[rgb]{0.906,0.969,0.973}}\textbf{69.06} & {\cellcolor[rgb]{0.906,0.969,0.973}}65.40          & {\cellcolor[rgb]{0.906,0.969,0.973}}\textbf{66.92} & {\cellcolor[rgb]{0.906,0.969,0.973}}\textbf{39.93} & {\cellcolor[rgb]{0.906,0.969,0.973}}40.80  \\
\bottomrule

\end{tabular}
}
\caption{Comparative results of structured pruning on LLaMA-7B at 20\% pruning ratio.The best results are shown in bold. And the proposed AdaPruner outperformed the baselines significantly (p \textless 0.05).}
\label{tab:Table_1}
\end{table*}

\section{Experiments}
\subsection{Datasets}
To evaluate the performance of LLMs before and after pruning, we use the perplexity (PPL) metric on the WikiText2~\cite{Merity2017} and PTB~\cite{Marcus1993} datasets to measure the language modeling capability. To evaluate the performance of the pruning method comprehensively and intuitively, we evaluate the zero-shot performance on seven commonsense reasoning datasets, including BoolQ~\cite{Clark2019}, PIQA~\cite{Bisk2020}, HellaSwag~\cite{Zellers2020}, WinoGrande~\cite{Sakaguchi2021}, ARC (including ARC-easy and ARC-challenge)~\cite{Clark2018}, and OpenbookQA~\cite{Mihaylov2018}. We report the accuracy of each dataset and the overall average accuracy across all datasets. A brief description of each dataset is shown in Appendix B.

\subsection{Baselines}
We comprehensively compare AdaPruner with existing structured pruning methods for LLMs. To ensure fairness in the comparison, all baselines use the same pruning and fine-tuning data set as LLM-Pruner, and 10 samples are selected as the calibration data. The baselines include Magnitude~\cite{Han2016}, Wanda~\cite{Sun2023}, LLM-Pruner~\cite{Ma2023}, LoRAShear~\cite{Chen2023} and MoreauPruner~\cite{Wang2024}. The details of the baseline models are described in Appendix B.
\begin{figure}[!t]
    \centering
    \includegraphics[scale=0.25]{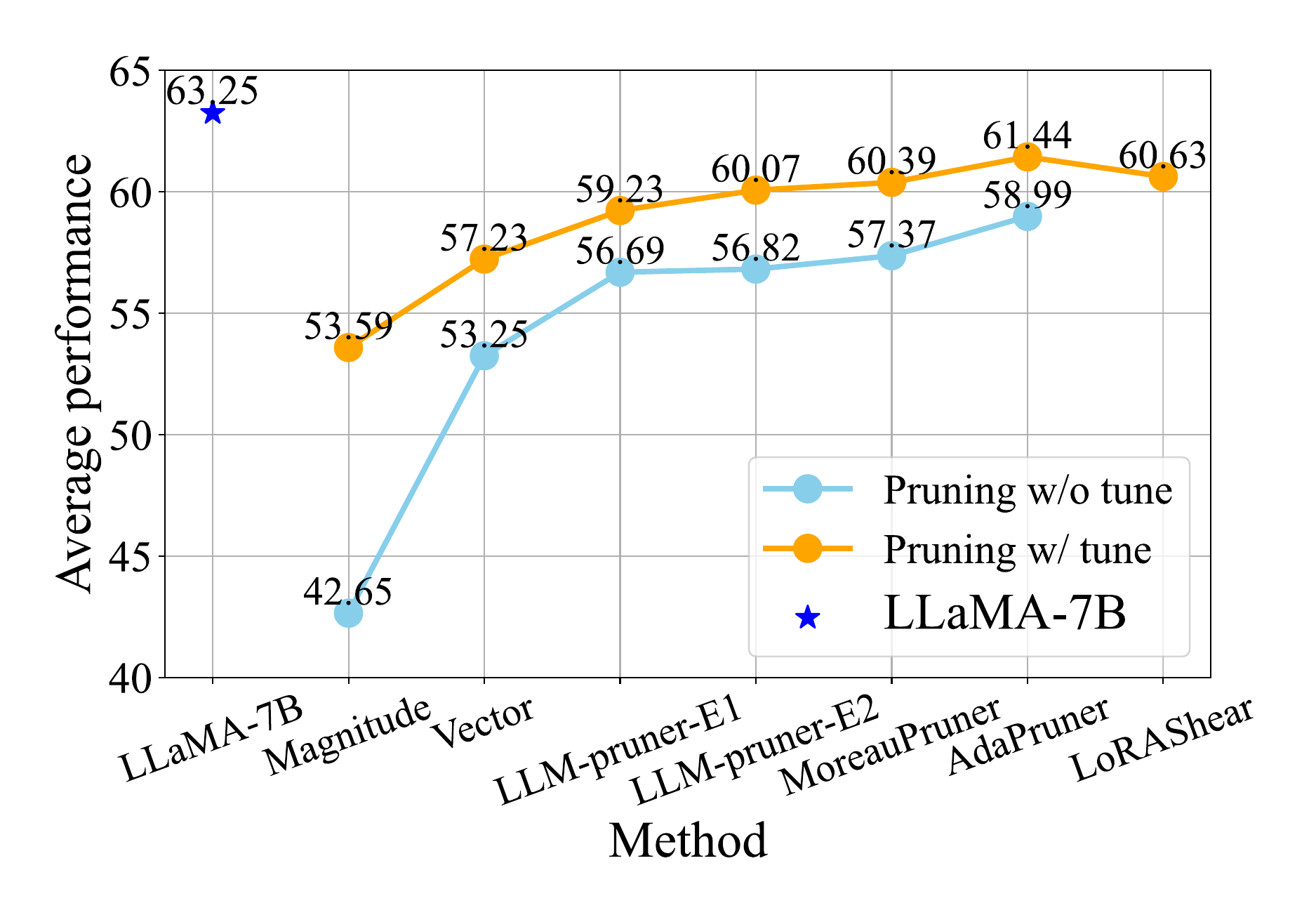}
    \caption{Average performance of structured pruning on LLa-MA-7B at 20\% pruning ratio.} 
    \label{fig:Figure_3} 
\end{figure}

\subsection{Implementation Details}
To evaluate the effectiveness of AdaPrune, we conduct experiments on the LLaMA-7B~\cite{Touvron2023} and Vicuna-7B~\cite{chiang2023vicuna} models. Meanwhile, samples with lengths less than 128 in BookCorpus~\cite{Zhu2015} are eliminated to narrow the calibration data subspace. The optimization range of the balance coefficients $\alpha_1$ and $\alpha_2$ for coarse and fine-grained importance estimation is set between 0 and 1. The estimation metrics alignment factors are optimized in the range {(e5, e6, e7), (e-2, e-3), (1e-4)}. Both related works and detailed experimental settings are in Appendix A and B.
\subsection{Comparative Results and Analysis}
We evaluate the zero-shot capabilities of AdaPruner and baselines on 7 downstream commonsense reasoning tasks, as well as language modeling on the WikiText2 and PTB datasets. We performed 20\% parameter pruning ratio experiments on LLaMA-7B, as shown in Table~\ref{tab:Table_1} and~\ref{fig:Figure_3}. The model pruned with AdaPruner exhibits the best average performance on commonsense reasoning benchmarks while retaining most language modeling capabilities. Specifically, the model the pruned model still retained 93.3\% of the average zero-shot performance of the unpruned model without fine-tuning. LLM-Pruner retained 89.8\% of the original model's performance, and AdaPruner improved by 3.5\%. Notably, AdaPruner significantly outperforms existing structured pruning methods without fine-tuning, suggesting that AdaPruner can obtain a more efficient pruning structure.

The pruned LLaMA-7B retains 97\% performance on the zero-shot evaluation dataset compared to the unpruned model with fine-tuning. Meanwhile, AdaPruner consistently outperforms existing structured pruning techniques, achieving 61.44\% average accuracy. In addition, it outperforms LoRAShear and MoreauPruner on the most zero-shot commonsense reasoning tasks, resulting in the best average performance among all baseline methods. Specifically, AdaPruner outperforms LLM-Pruner-E2, LoRAShear, and MoreauPruner by 1.37\%, 0.81\%, and 1.05\% on average performance, respectively. AdaPruner outperforms the other baseline methods in preserving text generation capabilities, further demonstrating the effectiveness of the AdaPruner method in preserving the language modeling ability and its superiority over existing structured pruning methods.
\begin{table*}
\centering
\setlength{\extrarowheight}{0pt}
\addtolength{\extrarowheight}{\aboverulesep}
\addtolength{\extrarowheight}{\belowrulesep}
\setlength{\aboverulesep}{0pt}
\setlength{\belowrulesep}{0pt}

\begin{tabular}{c!{\vrule width \lightrulewidth}ccccccc!{\vrule width \lightrulewidth}c} 
\toprule
Method                & BoolQ$\uparrow$                                   & PIQA$\uparrow$                                    & HellaSwag$\uparrow$                               & WinoGrande$\uparrow$                              & ARC-e$\uparrow$                                   & ARC-c$\uparrow$                                   & OBQA$\uparrow$                                    & Average$\uparrow$                                  \\ 
\midrule
AdaPruner             & {\cellcolor[rgb]{0.906,0.969,0.973}}\textbf{70.34} & {\cellcolor[rgb]{0.906,0.969,0.973}}\textbf{77.69} & {\cellcolor[rgb]{0.906,0.969,0.973}}\textbf{69.06} & {\cellcolor[rgb]{0.906,0.969,0.973}}\textbf{65.40} & {\cellcolor[rgb]{0.906,0.969,0.973}}\textbf{66.92} & {\cellcolor[rgb]{0.906,0.969,0.973}}\textbf{39.93} & {\cellcolor[rgb]{0.906,0.969,0.973}}\textbf{40.80} & {\cellcolor[rgb]{0.906,0.969,0.973}}\textbf{61.44}  \\ 
\midrule
\multicolumn{9}{c}{\textit{Importance Estimation Metrics }}                                                                                                                                                                                                                                                                                                                                                                                            \\ 
\midrule
AdaPruner w/o $I_1$   & 70.00                                             & 76.61                                             & 68.95                                             & 64.01                                             & 66.20                                             & 37.88                                             & 40.60                                             & 60.61                                              \\
AdaPruner w/o $I_2$   & 68.65                                             & 77.31                                             & 67.79                                             & 63.85                                             & 66.58                                             & 38.57                                             & 39.60                                             & 60.34                                              \\ 
\midrule
\multicolumn{9}{c}{\textit{Calibration Data }}                                                                                                                                                                                                                                                                                                                                                                                                         \\ 
\midrule
AdaPruner w/o $D$     & 69.63                                             & 77.86                                             & 68.95                                             & 64.56                                             & 65.03                                             & 38.48                                             & 40.60                                             & 60.73                                              \\ 
\midrule
\multicolumn{9}{c}{\textit{Optimization Order }}                                                                                                                                                                                                                                                                                                                                                                                                       \\ 
\midrule
AdaPruner w/o unified & 69.81                                             & 77.09                                             & 69.03                                             & 65.23                                             & 66.37                                             & 39.16                                             & 40.00                                             & 60.98                                              \\
\bottomrule
\end{tabular}
\caption{Ablation study for optimization of importance estimation metrics, calibration data, and different optimization orders.}
\label{tab:Table_2}
\end{table*}

\begin{figure}[!t]
	\centering
	\subfigure{
    
		\label{fig:Figure_4a}
		\includegraphics[scale=0.225]{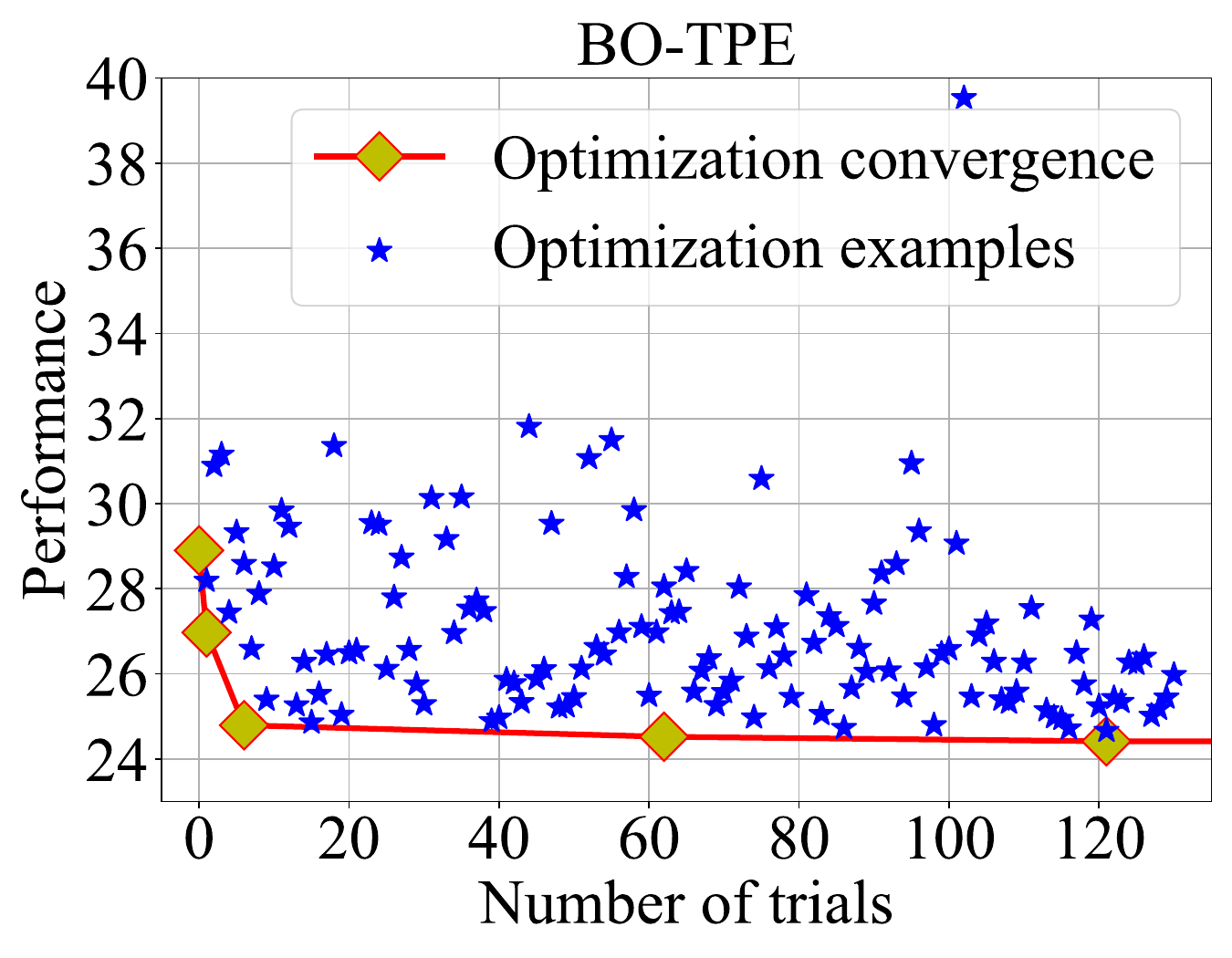}
	}
	\subfigure{
		\label{fig:Figure_4b}
		\includegraphics[scale=0.225]{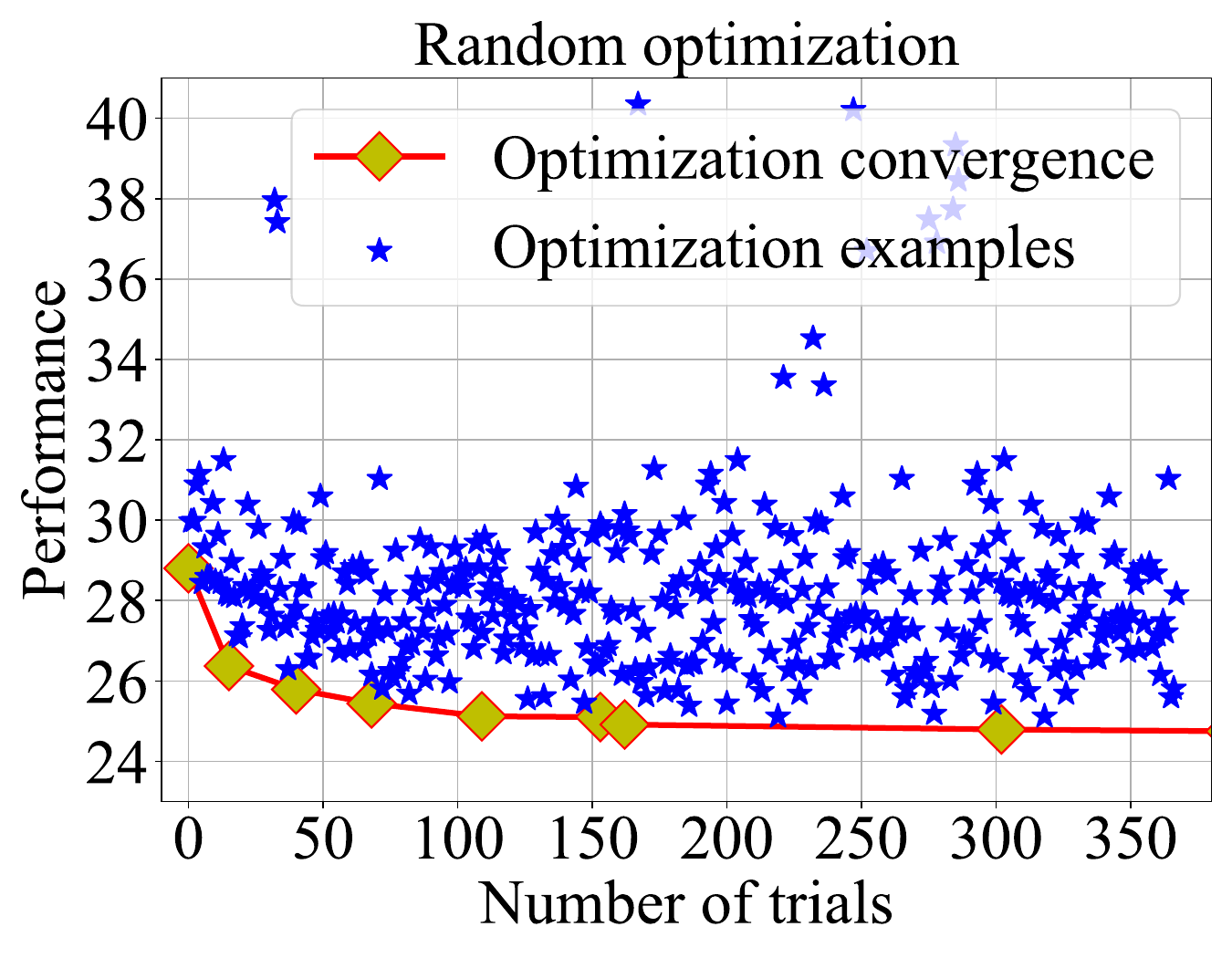}
	}
	\caption{Comparison of Bayesian and stochastic optimization processes.}
	\label{fig:Figure_4}
\end{figure}

\paragraph{Impact of calibration data.} To evaluate the importance of searching for calibration data, we removed the search for calibration data and replaced it with random selection (AdaPruner w/o $D$), As shown in Table~\ref{tab:Table_2}. The calibration data obtained using random selection shows a performance degradation of all downstream tasks. Specifically, the calibration data obtained with random selection performs less than that obtained with AdaPruner adaptive optimization by 0.71\% in average performance, demonstrating the validity of the calibration data adaptively obtained by AdaPruner and the importance of the calibration data in the pruning process. Notably, even when replacing randomized calibration data, the performance is still higher at 1.5\% and 0.66\% than that of LLM-Pruner-E1 and LLM-Pruner-E2, respectively, illustrating the robustness and validity of the searched importance estimation metrics under different calibration samples.
\begin{figure}[!t]
	\centering
	\subfigure{
		\label{fig:Figure_5a}
		\includegraphics[scale=0.248]{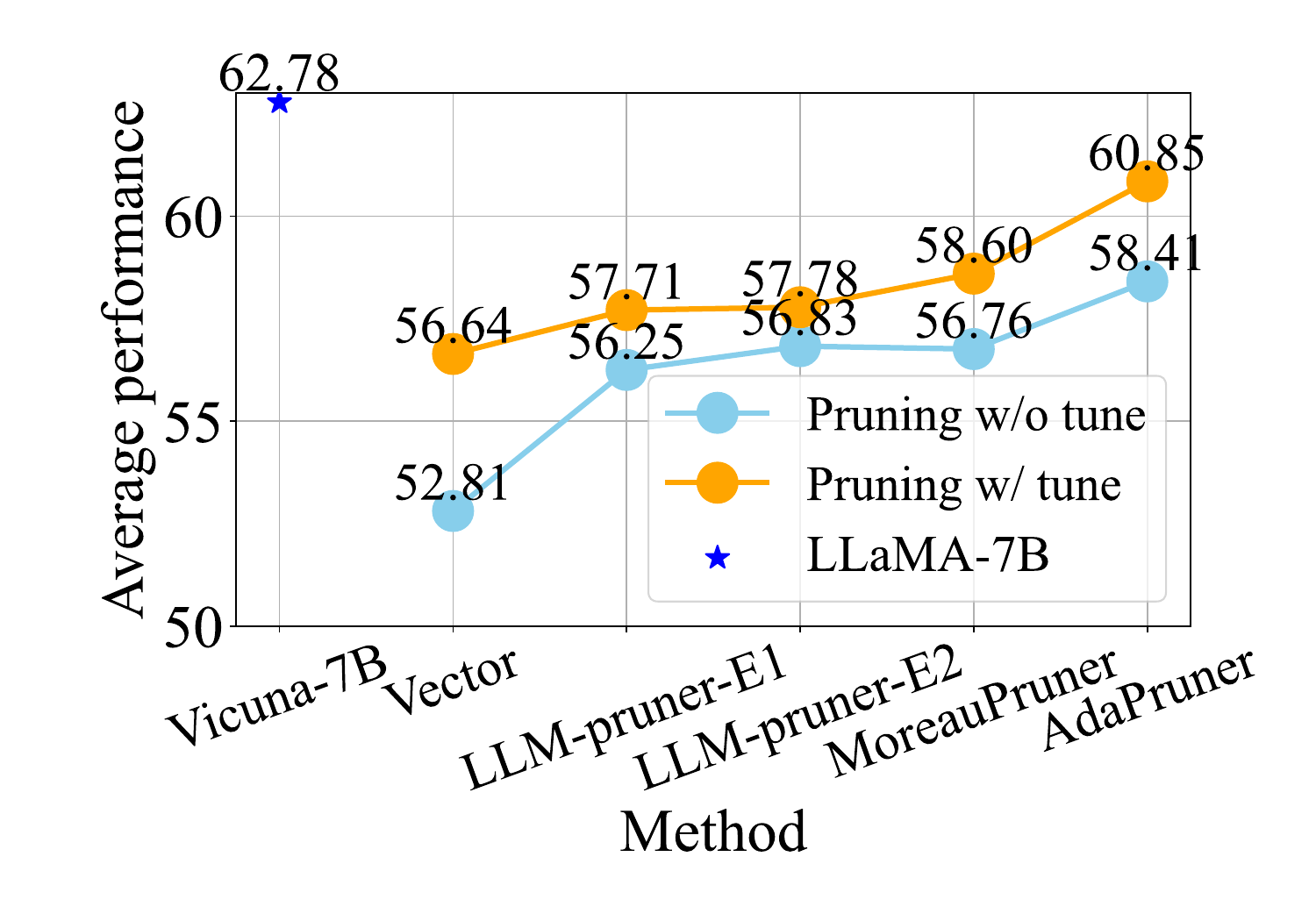}
	}
	\subfigure{
		\label{fig:Figure_5b}
		\includegraphics[scale=0.248]{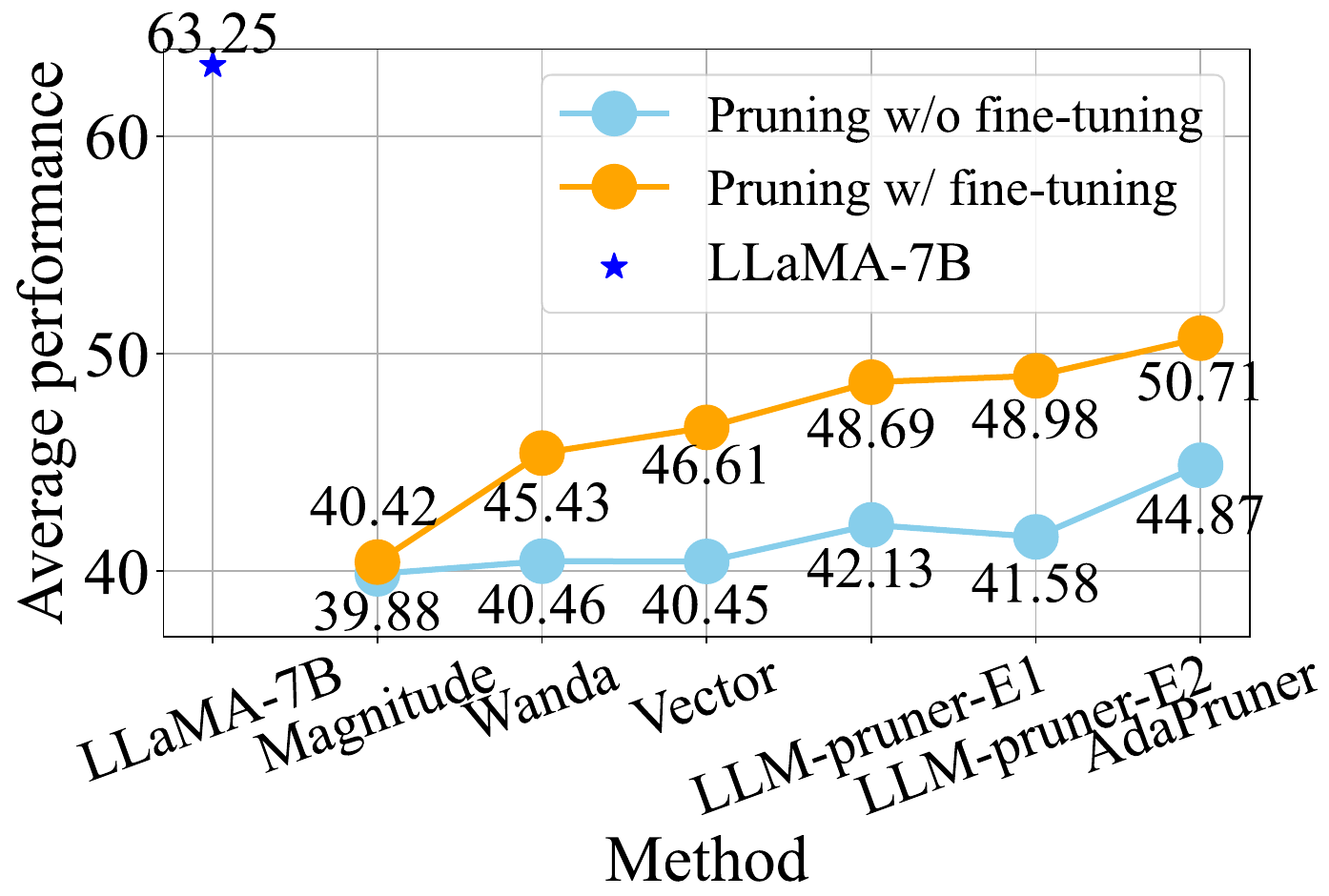}
	}
	\caption{Average performance of structured pruning for Vicuna-7B at 20\% pruning ratio and LLaMA-7B at 50\% pruning ratio.}
	\label{fig:Figure_5}
\end{figure}
\paragraph{Impact of joint optimization.} To demonstrate the effectiveness of end-to-end joint optimization, Table~\ref{tab:Table_2} compares different optimization order strategies. The joint optimization of calibration data and importance estimation metrics is replaced with a combined optimization (AdaPruner w/o unified). It can be observed that the combined optimization shows performance degradation on all tasks. The combined optimization underperforms the joint optimization by 0.46\% on the zero-shot average performance for commonsense reasoning. The reason is that the calibration data obtained by the combined optimization is not fully compatible with the importance estimation metrics, leading to inaccurate importance estimation and decreased model performance. This further proves that the joint optimization can find the globally optimal calibration data and importance estimation metrics, improving the pruned model's performance.
\begin{table*}
\centering
\setlength{\extrarowheight}{0pt}
\addtolength{\extrarowheight}{\aboverulesep}
\addtolength{\extrarowheight}{\belowrulesep}
\setlength{\aboverulesep}{0pt}
\setlength{\belowrulesep}{0pt}

\resizebox{\linewidth}{!}{
\begin{tabular}{cc|cc|ccccccc} 
\toprule
Pruning ratio                                                                & Method        & WikiText2$\downarrow$                              & PTB$\downarrow$                                    & BoolQ$\uparrow$                                    & PIQA$\uparrow$                                     & HellaSwag$\uparrow$                                & WinoGrande$\uparrow$                               & ARC-e$\uparrow$                                    & ARC-c$\uparrow$                                    & OBQA$\uparrow$                             \\ 
\midrule
Ratio=0\%                                                                    & Vicuna-7B     & 16.11                                              & 61.37                                              & 76.57                                              & 77.75                                              & 70.64                                              & 67.40                                              & 65.11                                              & 41.21                                              & 40.80                                      \\ 
\midrule
\multirow{5}{*}{\begin{tabular}[c]{@{}c@{}}Ratio=20\%\\w/ tune\end{tabular}} & Vector        & 19.94                                              & 74.66                                              & 63.15                                              & 74.59                                              & 61.95                                              & 60.30                                              & 60.48                                              & 36.60                                              & 39.40                                      \\
                                                                             & LLM-Pruner-E1 & 19.69                                              & 78.25                                              & 63.33                                              & 76.17                                              & 65.13                                              & 60.22                                              & 62.84                                              & 37.12                                              & 39.20                                      \\
                                                                             & LLM-Pruner-E2 & 18.97                                              & 76.78                                              & 60.40                                              & 75.63                                              & 65.45                                              & 63.22                                              & 63.05                                              & 37.71                                              & 39.00                                      \\
                                                                             & MoreauPruner  & 19.66                                              & 73.74                                              & 63.15                                              & \textbf{76.77}                                     & 65.96                                              & 60.85                                              & 65.74                                              & 37.12                                              & \textbf{40.60}                             \\
                                                                             & AdaPruner     & {\cellcolor[rgb]{0.906,0.969,0.973}}\textbf{18.05} & {\cellcolor[rgb]{0.906,0.969,0.973}}\textbf{73.56} & {\cellcolor[rgb]{0.906,0.969,0.973}}\textbf{71.93} & {\cellcolor[rgb]{0.906,0.969,0.973}}\textbf{76.77} & {\cellcolor[rgb]{0.906,0.969,0.973}}\textbf{67.06} & {\cellcolor[rgb]{0.906,0.969,0.973}}\textbf{64.80} & {\cellcolor[rgb]{0.906,0.969,0.973}}\textbf{66.12} & {\cellcolor[rgb]{0.906,0.969,0.973}}\textbf{40.27} & {\cellcolor[rgb]{0.906,0.969,0.973}}39.00  \\
\bottomrule
\end{tabular}
}
\caption{Comparative results of structured pruning on Vicuna-7B at 20\% pruning ratio with fine-tuning.}
\label{tab:Table_3}
\end{table*}

\begin{table*}
\centering
\setlength{\extrarowheight}{0pt}
\addtolength{\extrarowheight}{\aboverulesep}
\addtolength{\extrarowheight}{\belowrulesep}
\setlength{\aboverulesep}{0pt}
\setlength{\belowrulesep}{0pt}

\resizebox{\linewidth}{!}{
\begin{tabular}{cc|cc|ccccccc} 
\toprule
Pruning ratio                                                                & Method                                                   & WikiText2$\downarrow$                              & PTB$\downarrow$                                          & BoolQ$\uparrow$                                    & PIQA$\uparrow$                                     & HellaSwag$\uparrow$                                & WinoGrande$\uparrow$                               & ARC-e$\uparrow$                                    & ARC-c$\uparrow$                           & OBQA$\uparrow$                                      \\ 
\midrule
Ratio=0\%                                                                    & LLaMA-7B                                                 & 12.62                                              & 22.14                                                    & 73.18                                              & 78.35                                              & 72.99                                              & 67.01                                              & 67.45                                              & 41.38                                     & 42.40                                               \\ 
\midrule
\multirow{6}{*}{\begin{tabular}[c]{@{}c@{}}Ratio=50\%\\w/ tune\end{tabular}} & Magnitude                                                & 78.80                                              & 164.32                                                   & 47.40                                              & 54.36                                              & 33.49                                              & 53.10                                              & 37.88                                              & 26.60                                     & 30.12                                               \\
                                                                             & \multicolumn{1}{c!{\vrule width \lightrulewidth}}{Wanda} & 43.89                                              & \multicolumn{1}{c!{\vrule width \lightrulewidth}}{85.87} & 50.90                                              & 57.38                                              & 38.12                                              & 55.68                                              & 42.68                                              & \textbf{34.20}                            & 36.25                                               \\
                                                                             & Vector                                                   & 43.47                                              & 68.51                                                    & 61.11                                              & 64.96                                              & 40.52                                              & 51.54                                              & 46.38                                              & 28.33                                     & 32.40                                               \\
                                                                             & LLM-Pruner-E1                                            & 38.12                                              & 66.35                                                    & 60.28                                              & 69.31                                              & 47.06                                              & 53.43                                              & 45.96                                              & 29.18                                     & 35.60                                               \\
                                                                             & LLM-Pruner-E2                                            & 45.70                                              & 69.33                                                    & 61.47                                              & 68.82                                              & 47.56                                              & 55.09                                              & 46.46                                              & 28.24                                     & 35.20                                               \\
                                                                             & AdaPruner                                                & {\cellcolor[rgb]{0.906,0.969,0.973}}\textbf{34.29} & {\cellcolor[rgb]{0.906,0.969,0.973}}\textbf{53.40}       & {\cellcolor[rgb]{0.906,0.969,0.973}}\textbf{61.89} & {\cellcolor[rgb]{0.906,0.969,0.973}}\textbf{70.29} & {\cellcolor[rgb]{0.906,0.969,0.973}}\textbf{47.98} & {\cellcolor[rgb]{0.906,0.969,0.973}}\textbf{55.72} & {\cellcolor[rgb]{0.906,0.969,0.973}}\textbf{52.61} & {\cellcolor[rgb]{0.906,0.969,0.973}}30.12 & {\cellcolor[rgb]{0.906,0.969,0.973}}\textbf{36.40}  \\
\bottomrule
\end{tabular}
}
\caption{Comparative results of structured pruning on LLaMA-7B at 50\% pruning ratio with fine-tuning.}
\label{tab:Table_4}
\end{table*}
\subsection{Ablation Study}
\paragraph{Impact of importance estimation metrics.} We evaluate the impact of importance estimation metrics obtained after optimization on pruning. Specifically, we remove the optimization of importance estimation metrics and fix using first-order (AdaPruner w/o $I_1$) and second-order (AdaPruner w/o $I_2$ ) Taylor-expanded as alternatives. Table~\ref{tab:Table_2} shows the pruned model using first-order and second-order Taylor-expanded. It can be observed that using different importance estimation metrics exhibits different degrees of performance degradation on all the tasks. AdaPruner outperforms the first-order approach regarding average accuracy on the zero-shot of commonsense inference by 0.83\% and the second-order approach by 1.1\%. Since AdaPruner finds more appropriate and comprehensive importance estimation metrics in the solution space while balancing coarse and fine-grained importance estimation metrics. The results show that optimizing importance estimation metrics is crucial in improving performance. 

\paragraph{Impact of search algorithms.} To evaluate the effectiveness of Bayesian optimization, we use random search to replace Bayesian optimization, as shown in Figure~\ref{fig:Figure_4}. As can be seen from the figure, Bayesian optimization achieves faster convergence and better final search results compared to random search. Specifically, Bayesian optimization achieves convergence in less than 150 iterations, while random search requires about 300, showing the superiority of Bayesian optimization in terms of search efficiency and effectiveness.

\subsection{Effects of Different Pruned Models}
To validate the effectiveness and robustness of the AdaPruner method for pruning different models, we perform structured pruning of Vicuna-7B using the optimal calibration data and importance estimation obtained after optimization, as shown in Table~\ref{tab:Table_3} and Figure~\ref{fig:Figure_5}. The Vicuna-7B model is pruned at a pruning ratio 20\% with fine-tuning. The table shows that the zero-shot average performance of AdaPruner consistently outperforms that of Vector, LLM-Pruner, and MoreauPruner. AdaPruner achieves 60.85\% average performance, which is 3.07\% and 2.25\% higher than that of LLM-pruner-E2 and MoreauPruner, respectively, allowing the AdaPruner method can be applied to other LLMs while maintaining higher accuracy compared to existing structured pruning methods. These results also demonstrate the generalization ability and stability of the AdaPruner method for pruning different models. More results without fine-tuning are in Appendix C.

\subsection{Effects of Different Pruning Ratios}
To further analyze the performance and impact of AdaPruner under different pruning ratios, we conduct a 50\% pruning ratio experiment on the LLaMA-7B model with fine-tuning, as shown in Tables~\ref{tab:Table_4} and Figure~\ref{fig:Figure_5}. It shows that AdaPruner's model performance consistently outperforms LLM-Pruner and other baseline methods even when the pruning ratio is increased to 50\%. The AdaPruner outperforms LLM-Pruner in terms of perplexity on the WikiText2 and PTB datasets. Specifically, with a 50\% pruning ratio and no fine-tuning, AdaPruner's perplexity on Wik-iText2 and PTB is 34.29 and 53.40, respectively, which are 11.41 and 15.93 lower than LLM-pruner-E2. In addition, AdaPruner achieves an average performance of 50.71\% on seven zero-shot inference tasks, which is 1.73\% higher than LLM-pruner-E2. It highlights the performance advantage of AdaPruner at high pruning ratios and confirms its performance stability when the pruning ratio increases. 

\section{Conclusions}
This study proposes AdaPruner, a sample-aware adaptive structured pruning framework for LLMs, aiming to adaptively search for the optimal calibration data and importance estimation metrics to improve pruned LLMs' computational efficiency and performance. The experimental results show that AdaPruner maintains 97\% of the performance over the unpruned model and demonstrate the effectiveness of the proposed method for pruning LLMs. In addition, AdaPruner's generalization ability and robustness for different models and pruning ratios. Future work will explore the potential of AdaPruner for a broader range of pruning solution space designs and model types.

\section{Acknowledgments}
This work was supported by the National Natural Science Foundation of China (NSFC) under Grant Nos.61966038 and 62266051, and the Yunnan
Provincial Department of Education Science Foundation under Grant No.2024Y030. The authors would like to thank the anonymous reviewers for their constructive comments. 

\bibliography{References}

\newpage\newpage
\appendix
\section{Appendix}
\subsection{A. Related Work}
Various model compression techniques have emerged to address the large model sizes and high computation costs of deploying LLMs, including knowledge distillation, model pruning, and model quantization. 
\paragraph{Knowledge Distillation.} Knowledge distillation (KD) compresses LLMs by training a smaller student model to imitate the knowledge of an LLM that serves as the teacher model so that the student model can still maintain a high performance like that of the teacher model. KD for LLMs can be classified into white-box and black-box methods depending on whether they consider the internal structure of the teacher model. 
\begin{itemize}

\item \textbf{Black-box KD.} Techniques such as in-context learning~\cite{Dong2022} and chain-of-thought~\cite{Wei2022} can enable the student model to learn the different capabilities of the LLM thoroughly. In Meta-ICL~\cite{Min2022} and Metal-ICL~\cite{Chen2022}, LLMs are meta-trained on different tasks using contextual learning objectives and then fine-tuned on unseen tasks through contextual learning. SOCRATIC COT~\cite{Shridhar2023} distills the chains of through ability from LLMs. 

\item \textbf{White-box KD.} LLMs' parameters or internal logic can also be used in the distillation process. For example, since a text generation task's output space is more complex than a classification task, traditional KD objectives, such as Kullback-Leibler Divergence (KLD), are unsuitable for open text generation tasks. To address this issue, MiniLLM~\cite{Gu2024} minimizes the inverse KLD via a policy gradient technique to prevent the student model from overestimating the low probability region of the teacher distribution. To exploit the rich semantic and syntactic knowledge in the middle layer of the teacher's model, TED~\cite{Liang2023} designs a hierarchical distillation of task awareness, which aligns the student's hidden representations with the teacher's hidden representations at each layer to distill knowledge relevant to the target task selectively.
\end{itemize}
\paragraph{Model Quantization} Model quantization compresses LLMs by converting high-precision data types for model weights or activations to low-precision data types (e.g., 32-bit floating point numbers to 8-bit integers). The quantization methods can be divided into quantization-aware training and post-training quantization according to the necessity of retraining. 
\begin{itemize}

\item \textbf{Quantization-aware training (QAT).} QAT requires retraining the whole model with high computational overhead. PEQA~\cite{Kim2023} performs a two-stage process. First, the parameter matrices of each fully connected layer are quantized into low-bit integer matrices and scalar vectors. Subsequently, the scalar vectors are fine-tuned for each downstream task.

\item \textbf{Post-training quantization (PTQ).} PTQ is a low-cost quantization scheme and is, therefore, commonly used to compress LLMs. AWQ~\cite{Lin2023}  protects 1\% of the significant weights to considerably reduce the quantization error since the weights are not equally heavy. The activation-aware approach considers the importance of weight corresponding to more significant activation. Further, SpQR~\cite{Compression2023} identifies and separates the anomalous weights to store the weird weights with higher precision and compresses all other weights to 3-4 bits.
\end{itemize}
\paragraph{Model Pruning.} Model pruning aims to identify and remove unimportant components, such as neurons and layers in LLMs, to generate compact and accurate LLMs. The existing pruning techniques are broadly categorized into structured and unstructured pruning. 
\begin{itemize}

\item \textbf{Structured pruning.} It performs pruning operations on weight matrices, layers, or other meaningful structures, thereby preserving the functionality of the remaining LLMs and enabling efficient computation, providing a more hardware-friendly solution that reduces storage requirements and increases inference speed. For example, LLM-Pruner~\cite{Ma2023} is the first attempt at structured pruning of LLMs, reducing model computation and memory usage while keeping the overall structure of the LLMs. Specifically, it utilizes a dependency detection algorithm to identify coupled structures in the LLMs, scores the importance of the coupled structures, and selectively removes non-critical structures according to first-order and approximate Hessian-based information. Finally, LLM-Pruner uses Low-Rank-Adaptor (LoRA) for fine-tuning to recover the pruned model weights. Further, LoRAShear~\cite{Chen2023} utilizes the Lora Half-Space Projected Gradient (LHSPG) technique for incremental structured pruning and knowledge recovery. LoRAShear can be applied to various LLMs through dependency graph analysis and sparsity optimization. Recently, Sheared-LLaMA~\cite{Xia2024} is a directed structured pruning that prunes a large model to a specified target structure. It then trains the pruned model using dynamically loaded data based on the loss reduction ratio in each domain, thus improving data usage efficiency and model performance. However, Sheared-LLaMA uses substantial computational resources for subsequent pre-training to performance recovery. Like LLM-Pruner, the number of layers is constant. Shortened LLaMA~\cite{Kim2024} proposes that a simple deep pruning method can achieve promising performance and improve inference speed in a zero-shot task.

\begin{table*}
\centering

\resizebox{\linewidth}{!}{
\begin{tabular}{cl} 
\toprule
Dataset    & Description                                                                                                           \\ 
\midrule
WikiText2  & Predicting the next word in
a sentence                                                                                \\
PTB        & Understand grammatical
relationships within sentences                                                                 \\
BoolQ      & Understand grammatical
relationships within sentences                                                                 \\
PIQA       & Evaluating the model’s understanding
for the laws of the physical world                                               \\
HellaSwag  & Evaluating the model’s commonsense reasoning
  ability on natural language inference tasks.                           \\
WinoGrande & Evaluating the model’s
understanding of gender-related information.                                                   \\
ARC        & Evaluating the model’s
understanding of commonsense reasoning tasks, involving subsets of ARC-easy and
ARC-challenge  \\
OpenbookQA & Evaluating the model’s understanding on
  question-answer tasks                                                       \\
\bottomrule
\end{tabular}
}
\caption{Details of datasets.}
\label{tab:Table_5}
\end{table*}

\begin{table*}
\centering
\setlength{\extrarowheight}{0pt}
\addtolength{\extrarowheight}{\aboverulesep}
\addtolength{\extrarowheight}{\belowrulesep}
\setlength{\aboverulesep}{0pt}
\setlength{\belowrulesep}{0pt}

\begin{tabular}{cc|ccccccc} 
\toprule
Pruning ratio                                                                & Method                                                           & BoolQ$\uparrow$                           & PIQA$\uparrow$                                     & HellaSwag$\uparrow$                                & WinoGrande$\uparrow$                               & ARC-e$\uparrow$                                    & ARC-c$\uparrow$                                    & OBQA$\uparrow$                                      \\ 
\midrule
Ratio=0\%                                                                    & Vicuna-7B                                                        & 76.57                                     & 77.75                                              & 70.64                                              & 67.40                                              & 65.11                                              & 41.21                                              & 40.80                                               \\ 
\midrule
\multirow{5}{*}{\begin{tabular}[c]{@{}c@{}}Ratio=20\%\\w/o tune\end{tabular}} & Vector                                                           & 62.17                                     & 71.44                                              & 55.80                                              & 53.43                                              & 55.77                                              & 33.28                                              & 37.80                                               \\
                                                                             & LLM-Pruner-E1                                                    & 61.70                                     & 75.30                                              & 63.75                                              & 56.20                                              & 63.22                                              & 36.60                                              & 37.00                                               \\
                                                                             & \multicolumn{1}{c!{\vrule width \lightrulewidth}}{LLM-Pruner-E2} & \textbf{62.87}                            & 75.41                                              & 64.00                                              & 58.41                                              & 60.98                                              & 37.12                                              & 39.00                                               \\
                                                                             & MoreauPruner                                                     & 56.82                                     & 75.79                                              & 64.73                                              & 56.35                                              & \textbf{65.95}                                              & 37.88                                              & 39.80                                               \\
                                                                             & AdaPruner                                                        & {\cellcolor[rgb]{0.906,0.969,0.973}}61.92 & {\cellcolor[rgb]{0.906,0.969,0.973}}\textbf{75.81} & {\cellcolor[rgb]{0.906,0.969,0.973}}\textbf{65.96} & {\cellcolor[rgb]{0.906,0.969,0.973}}\textbf{61.33} & {\cellcolor[rgb]{0.906,0.969,0.973}}64.18 & {\cellcolor[rgb]{0.906,0.969,0.973}}\textbf{39.68} & {\cellcolor[rgb]{0.906,0.969,0.973}}\textbf{40.00}  \\
\bottomrule
\end{tabular}
\caption{Comparative results of structured pruning on Vicuna-7B at 20\% pruning ratio without fine-tuning.}
\label{tab:Table_6}
\end{table*}
\begin{table*}[!ht]
\centering
\setlength{\extrarowheight}{0pt}
\addtolength{\extrarowheight}{\aboverulesep}
\addtolength{\extrarowheight}{\belowrulesep}
\setlength{\aboverulesep}{0pt}
\setlength{\belowrulesep}{0pt}

\begin{tabular}{cc|ccccccc} 
\toprule
Pruning ratio                                                                & Method                                                           & BoolQ$\uparrow$                                    & PIQA$\uparrow$                                     & HellaSwag$\uparrow$                                & WinoGrande$\uparrow$                      & ARC-e$\uparrow$                                             & ARC-c$\uparrow$                           & OBQA$\uparrow$                                      \\ 
\midrule
Ratio=0\%                                                                    & LLaMA-7B                                                         & 73.18                                              & 78.35                                              & 72.99                                              & 67.01                                     & 67.45                                                       & 41.38                                     & 42.40                                               \\ 
\midrule
\multirow{6}{*}{\begin{tabular}[c]{@{}c@{}}Ratio=50\%\\w/o tune\end{tabular}} & Magnitude                                                        & 44.10                                              & 54.98                                              & 31.27                                              & 52.93                                     & 38.76                                                       & \textbf{27.50}                            & 29.67                                               \\
                                                                             & \multicolumn{1}{c!{\vrule width \lightrulewidth}}{Wanda}         & 45.13                                              & 55.54                                              & 31.37                                              & \textbf{55.87}                            & 39.43                                              & 25.76                                     & 30.12                                               \\
                                                                             & Vector                                                           & 60.17                                              & 55.11                                              & 27.25                                              & 49.88                                     & 29.00                                                       & 25.77                                     & 34.00                                               \\
                                                                             & \multicolumn{1}{c!{\vrule width \lightrulewidth}}{LLM-Pruner-E1} & 52.32                                              & 59.63                                              & 35.64                                              & 53.20                                     & 33.50                                                       & 27.22                                     & 33.40                                               \\
                                                                             & LLM-Pruner-E2                                                    & 52.57                                              & 60.45                                              & 35.86                                              & 49.01                                     & 32.83                                                       & 25.51                                     & 34.80                                               \\
                                                                             & AdaPruner                                                        & {\cellcolor[rgb]{0.906,0.969,0.973}}\textbf{60.31} & {\cellcolor[rgb]{0.906,0.969,0.973}}\textbf{63.11} & {\cellcolor[rgb]{0.906,0.969,0.973}}\textbf{37.74} & {\cellcolor[rgb]{0.906,0.969,0.973}}53.85 & {\cellcolor[rgb]{0.906,0.969,0.973}}\textbf{39.77} & {\cellcolor[rgb]{0.906,0.969,0.973}}26.11 & {\cellcolor[rgb]{0.906,0.969,0.973}}\textbf{37.20}  \\
\bottomrule
\end{tabular}
\caption{Comparative results of structured pruning on LLaMA-7B at 50\% pruning ratio without fine-tuning.}
\label{tab:Table_7}
\end{table*}

\item \textbf{Unstructured pruning.} It performs pruning operations on weights or neurons instead of structured blocks. Although unstructured pruning typically results in higher compression ratios, the need for specific hardware support leads to the failure to achieve proper inference acceleration or storage reduction. SparseGPT~\cite{Frantar2023} is the first approach to perform unstructured pruning on LLMs. It applies the Hessian inverse to pruning and uses a complex weight-updating process that includes synchronized second-order Hessian updates. The strategy is computationally expensive to execute iteratively between weight pruning and weight updating at each layer. In contrast, Wanda~\cite{Sun2023} removes unimportant weights by introducing a new criterion of combining the weight size with its input activation to retain anomalous features without updating the remaining weights. Both can be extended to semi-structured pruning (i.e., $N$: $M$ sparsity).
\end{itemize}
\begin{table*}[!ht]
\centering
\setlength{\extrarowheight}{0pt}
\addtolength{\extrarowheight}{\aboverulesep}
\addtolength{\extrarowheight}{\belowrulesep}
\setlength{\aboverulesep}{0pt}
\setlength{\belowrulesep}{0pt}

\begin{tabular}{c|cccccccc} 
\toprule
Method                                                       & BoolQ$\uparrow$                                   & PIQA$\uparrow$                                    & HellaSwag$\uparrow$                               & WinoGrande$\uparrow$                              & ARC-e$\uparrow$                                   & ARC-c$\uparrow$                                   & OBQA$\uparrow$                                    & Average$\uparrow$                                  \\ 
\midrule
LLaMA-7B                                                     & 73.18                                             & 78.35                                             & 72.99                                             & 67.01                                             & 67.45                                             & 41.38                                             & 42.40                                             & 63.25                                              \\ 
\midrule
\multicolumn{1}{c!{\vrule width \lightrulewidth}}{AdaPruner} & {\cellcolor[rgb]{0.906,0.969,0.973}}\textbf{70.34} & {\cellcolor[rgb]{0.906,0.969,0.973}}\textbf{77.69} & {\cellcolor[rgb]{0.906,0.969,0.973}}\textbf{69.06} & {\cellcolor[rgb]{0.906,0.969,0.973}}\textbf{65.40} & {\cellcolor[rgb]{0.906,0.969,0.973}}\textbf{66.92} & {\cellcolor[rgb]{0.906,0.969,0.973}}\textbf{39.93} & {\cellcolor[rgb]{0.906,0.969,0.973}}\textbf{40.80} & {\cellcolor[rgb]{0.906,0.969,0.973}}\textbf{61.44}  \\
w/ 10 space $D$                                              & 69.72                                             & 77.15                                             & 68.76                                             & 64.33                                             & 64.02                                             & 39.16                                             & 40.40                                             & 60.51                                              \\
w/ score $D$                                                 & 67.49                                             & 76.93                                             & 68.25                                             & 64.64                                             & 66.58                                             & 38.14                                             & 40.60                                             & 60.38                                              \\
\bottomrule
\end{tabular}
\caption{Experimental results based on different ways of constructing calibration data subspaces.}
\label{tab:Table_8}
\end{table*}

\begin{table*}[!ht]
\centering

\begin{tabular}{c!{\vrule width \lightrulewidth}cccc} 
\toprule
Method    & \#Parameters & \#Memory & \#Operator & \#Delay  \\ 
\midrule
LLaMA-7B  & 6.74B        & 425.12G  & 12892.6MiB & 4.93s    \\
AdaPruner & 5.42B        & 340.48G  & 10383.6MiB & 2.40s    \\
\bottomrule
\end{tabular}
\caption{Statistical information on model pruning.}
\label{tab:Table_9}
\end{table*}

\subsection{B. Detailed Experimental Settings}

\paragraph{Datasets and Metrics.} The experiments are conducted on WikiText2 and PTB datasets. The PPL is used as the evaluation metric, where lower perplexity indicates that the language model is more predictive of a given text sequence. We report the accuracy of each dataset and the overall average accuracy across all commonsense reasoning datasets. Table~\ref{tab:Table_5} shows a brief description of each dataset.

\noindent \textbf{Baselines. }    All baseline models are described as follows:
\begin{itemize}
    \item \textbf{Magnitude.} The importance of model weights is assessed based on the absolute values of the weight matrix, and weights with smaller absolute values are less informative.
    \item \textbf{Wanda.} The importance of model weights is assessed by multi-plying the weight magnitude with the corresponding input activation.
    \item \textbf{LLM-Pruner.} The importance of model weights is assessed by calculating the error caused by removing the weights. As a result, it requires the computation of gradient information to selectively remove non-critical coupling structures from LLMs and perform post-pruning LoRA fine-tuning. Vector denotes the structured pruning of LLMs using the importance assessment metrics in Eq.~\eqref{eq:Equation_2}. LLM-Pruner-E denotes the importance assessment using the $n$th-order term in the Taylor expansion of Eq.~\eqref{eq:Equation_4} to achieve structured pruning. 
    \item \textbf{LoRAShear.} A dependency graph is built for the model weights to find the minimal removal structure, and the weights are gradually pruned according to the LoRA half-space projection gradient. 
    \item \textbf{MoreauPruner.} A weight perturbation structure pruning method is introduced, which is robust to weight perturbations.
\end{itemize}

\begin{table*}
\centering

\resizebox{\linewidth}{!}{
\begin{tabular}{c!{\vrule width \lightrulewidth}l} 
\toprule
Model      & \multicolumn{1}{c}{Generation sentence}                                                                                                                                                                                                                                                                                                                                                                                                                                                   \\ 
\midrule
LLaMA-7B   & \begin{tabular}[c]{@{}l@{}}\textbf{The universe is the entity of space, time, matter, and energy that exists}. It includes all matter and energy, whether visible~\\ornot, and in fact the whole universe is invisible. There are many theories regarding its origin and ultimate fat…\end{tabular}                                                                                                                                                                                       \\ 
\midrule
LLM-Pruner & \begin{tabular}[c]{@{}l@{}}\textbf{The universe is the entirety of space, time, matter, and energy that exists.} The laws that make up physics — electrons \\orbiting atoms, magnets attracting other objects, planets orbiting the Sun—have been consistent over…\end{tabular}                                                                                                                                                                                                           \\ 
\midrule
AdaPruner  & \begin{tabular}[c]{@{}l@{}}\textbf{The universe is the entity of space, time, matter, and energy that exists. }This includes things we can’t see, such as atoms, \\electrons, particles, the vacuum, and the cosmic background radiation. Theoretically, the universe can contain anything, from \\planets, stars, galaxies, black holes and quantum fields. So far, the most widely accepted theory about the origin and evolution \\of the universe is the Big Bang theory…\end{tabular}  \\
\bottomrule
\end{tabular}
}
\caption{Comparison of generation sentences.}
\label{tab:Table_10}
\end{table*}

\paragraph{Implementation Details.} During the pruning process, 10 high-quality samples are adaptively selected from the BookCorpus dataset as the calibration data for calculating the gradient of the models, which in turn assesses the importance of different structures. The sequence length is fixed at 128, and the pruning ranges from the 4-th layer to the 29-th layer. AdaPruner has fine-tuned 2 epochs on the Alpaca dataset~\cite{taori2023stanford}. The batch size is 128. The learning rate is 1e-4, and the model is optimized using the AdamW optimizer. For the 7B LLMs, a single pruning takes 5 minutes on a single NVIDIA RTX 3090 GPU. 

\subsection{C. Additional Experiments}
The pruning ratio refers to the ratio of the total number of pruned parameters to the total number of unpruned model parameters.
\paragraph{More Results of Different Pruned Models.} The Vicuna-7B model is pruned at a pruning ratio of 20\% without fine-tuning. The pruned model is evaluated using the perplexity metrics on the WikiText2 and PTB datasets and the zero-shot accuracy of the seven commonsense inferences. The detailed results are shown in Table~\ref{tab:Table_6}. AdaPruner achieves better performance than other baseline models in most commonsense reasoning tasks. AdaPruner achieves better performance than the baseline model with and without fine-tuning, proving the effectiveness of AdaPruner.

\paragraph{More Results of Different Pruning Ratios.} To further analyze the performance and impact of AdaPruner under different pruning ratios, we conduct a 50\% pruning ratio experiment on the LLaMA-7B model without fine-tuning. Table~\ref{tab:Table_7} shows the accuracy of AdaPruner with multiple baseline models at a 50\% pruning ratio on seven zero-shot commonsense inference tasks, as well as the perplexity of the WikiText2 and PTB datasets. In particular, without fine-tuning, AdaPruner significantly outperforms LLM-Pruner regarding perplexity on the WikiText2 and PTB datasets. Specifically, with a 50\% pruning ratio and without fine-tuning, AdaPruner's perplexity on WikiText2 and PTB is 101.61 and 239.96, respectively, which are lower than 4.46 and 26.69 than LLM-pruner-E2. This highlights the performance advantage of AdaPruner at high pruning ratios.

\paragraph{Effects of Calibration Data Solution Space.} To better understand the importance of calibration data solution space design, we explore two approaches to constructing calibration data subspaces, as shown in Table~\ref{tab:Table_8}. We set up two experiments: (1) w/ 10 space D, which indicates that the calibration data subspace is replaced by dividing it into 10 solution subspaces. Subsequently, Bayesian optimization is used to search for the calibration data. (2) w/ score D denotes that each dataset sample is scored based on the average perplexity on the WikiText2 and PTB datasets, and the top 10 samples with the highest scores are selected as the calibration data. The results in Table 8 show that optimizing all calibration samples directly in the uniform solution space exhibits superior performance compared to the other two designs. Specifically, AdaPruner outperforms w/ 10 spac e D and w/ score D by 0.93\% and 1.06\%, respectively. The rationale is that w/ 10 space D optimizes in the space of 10 independent solutions, which is easy to select into the local optimum. Meanwhile, w/ score D neglects the diversity information after combining calibration samples, resulting in poor pruning performance. These findings powerfully demonstrate the effectiveness of the proposed solution space design for calibration data.

\paragraph{Statistical Information on Model Pruning.} Compared to unstructured pruning, structured pruning reduces the number of parameters and facilitates deployment without relying on specific hardware. Table~\ref{tab:Table_9} shows the key statistics of the pruning model, including the number of parameters, MACs, memory, and latency. The inference latency is derived from an inference mode test on the WikiText2 dataset using a single NVIDIA RTX 3090 GPU. The input to the model is a sentence with 64 lengths. With a pruning ratio of 20\%, AdaPruner reduces the number of parameters by 19.6\%, leading to a corresponding 20\% reduction in MACs. The latency of the pruned model is approximately two times faster than the original model. The AdaPruner method has achieved significant results in parameter pruning and latency optimization.

\paragraph{Case Study} Table~\ref{tab:Table_10} provides some sentences generated by LLaMA-7B, LLM-Pruner and AdaPruner with a pruning ratio of 20\%, respectively. The sentences generated by the AdaPruner pruning model are comparable to those generated by the original model. AdaPruner is like the unpruned LLaMA-7B model regarding fluency, relevance, and information content for a given topic. Notably, the AdaPruner pruned model outperforms LLM-Pruner regarding sentence coherence, logic, and content relevance.

\end{document}